\documentclass[conference]{IEEEtran}

\usepackage{amssymb}
\usepackage[english]{babel}
\usepackage{array}
\usepackage[ruled,section]{algorithm}
\usepackage[noend]{algorithmic}
\usepackage{multirow}
\usepackage{color}

\usepackage{cite}

\ifCLASSINFOpdf
   \usepackage[pdftex]{graphicx}
\else
   \usepackage[dvips]{graphicx}
\fi

\ifCLASSOPTIONcompsoc
    \usepackage[caption=false,font=normalsize,labelfont=sf,textfont=sf]{subfig}
\else
    \usepackage[caption=false,font=footnotesize]{subfig}
\fi

\usepackage[cmex10]{amsmath}
\allowdisplaybreaks

\usepackage{mdwmath}
\usepackage{mdwtab}
\usepackage{fixltx2e}
\usepackage{url}
\usepackage{epstopdf}

\begin{document}

\title{Dynamic Hierarchical Dirichlet Process for Abnormal Behaviour Detection in Video}

\author{\IEEEauthorblockN{Olga Isupova, Danil Kuzin, Lyudmila Mihaylova}
\IEEEauthorblockA{Department of Automatic Control and System Engineering,
University of Sheffield\\
Sheffield, UK\\
Email: o.isupova@sheffield.ac.uk, dkuzin1@sheffield.ac.uk, l.s.mihaylova@sheffield.ac.uk}}

\maketitle

\begin{abstract}
This paper proposes a novel dynamic Hierarchical Dirichlet Process topic model that considers the dependence between successive observations. Conventional posterior inference algorithms for this kind of models require processing of the whole data through several passes. It is computationally intractable for massive or sequential data. We design the batch and online inference algorithms, based on the Gibbs sampling, for the proposed model. It allows to process sequential data, incrementally updating the model by a new observation. The model is applied to abnormal behaviour detection in video sequences. A new abnormality measure is proposed for decision making. The proposed method is compared with the method based on the non-dynamic Hierarchical Dirichlet Process, for which we also derive the online Gibbs sampler and the abnormality measure. The results with synthetic and real data show that the consideration of the dynamics in a topic model improves the classification performance for abnormal behaviour detection.   
\end{abstract}

\IEEEpeerreviewmaketitle

\section{Introduction}

Unsupervised and semi-supervised learning for various video processing applications is an active research area nowadays. In many situations supervised learning is inappropriate or impossible. For example, in abnormal behaviour detection it is difficult to predict in advance what kind of abnormality may happen, collect and label a training dataset for some supervised learning algorithm. 

Within the unsupervised methods topic modeling is a promising approach for abnormal behaviour detection~\cite{Jeong14, Varadarajan2009, Mehran09}. It allows not only to give warnings about abnormalities but also provides an information about typical patterns of behaviour or motion. 

Topic modeling~\cite{Hofmann99, Blei03LDA} is a statistical tool for discovering a latent structure in data. In text mining it is assumed that unlabelled documents can be represented as mixtures of topics, where the topics are distributions over words. The topics are latent and the inference in topic models is aimed to discover them.

In the conventional topic models, documents are independent. They share the same set of topics, but weights in a topic mixture for a particular document are independent of weights for all other documents in a dataset. However, in some cases it is reasonable to assume dependence in topic mixtures in different documents. 

Consider the analysis of scientific papers of a given conference in text mining. It is expected that if a topic is ``hot'' in a given year, it would be popular in the next year too. The popularity of the topics changes through the years but in each two successive years the set of popular topics would be similar. It means that in a topic model the topic mixtures in the documents in successive years are similar to each other. 

The same ideas are valid for abnormal behaviour detection. Documents are usually defined as short video clips extracted from a whole video sequence. Topics represent some local motion patterns. If the clips are sufficiently short, motions started in a given clip would continue in the next clip. Therefore it may be expected that the topic mixtures in the successive clips would be similar. 

In this paper the dynamic topic model is proposed to improve the performance of abnormal behaviour detection. Two types of dynamics are considered in the topic modeling literature. In the first type the dynamics is assumed on the topic mixtures in documents~\cite{Hospedales2011, Kuettel2010, PruteanuMalinici2010}. This type of the dynamics is described earlier. In the second type the dynamics is assumed on the topics themselves~\cite{Wang08, Fu2016, Chen2012}, i.e. the distributions over words, which correspond to topics, change through time. There are works where both types of the dynamics are considered~\cite{Blei2006Dynamic, Ahmed2010}.

In the proposed model the first type of the dynamics is considered. The model is constructed to encourage neighbour documents to have similar topic mixtures. The second type of the dynamics is not assumed, as in the video processing the set of words and their popularity do not change, thus the distributions over words are not expected to change.

Imagine there is an infinitely long video sequence. Motion patterns, which are typical for a scene, may appear and disappear and the total number of these patterns may be infinite. The motion patterns are modelled as topics in the topic model, hence the number of topics in the topic model may potentially be infinite. This kind of intuition may be simulated by a nonparametric model~\cite{Orbanz2011}. Therefore the proposed model is nonparametric.

The most related model to the proposed one is presented in~\cite{Ahmed2010}, which is also a dynamic topic model. The main difference between this model and the proposed one is that in the later a document, although is encouraged to have a topic mixture similar to the one in the previous document, may have any of the topics used in the dataset so far.

In abnormal behaviour detection it is essential to make a decision as soon as possible to warn a human operator to react. 
We propose batch and online inference for the model based on the Gibbs sampler. During the batch offline set up the Gibbs sampler processes a training set of documents, estimating distributions of words in topics. During the online set up testing documents are processed one by one. The main goal of the online inference is to estimate a topic mixture for the current document, without reconsidering all the previous documents.
We also propose an abnormality measure, which is used in the final decision making. 

The rest of the paper is organised as follows. In section~\ref{sec:video_rep} visual words and documents are defined. The proposed model is described in section~\ref{sec:proposed_model}. Section~\ref{sec:inference} presents the inference for the model, while section~\ref{sec:abnormality} introduces the abnormality detection procedure. The experimental results are given in section~\ref{sec:experiments}. Section~\ref{sec:conclusions} concludes the paper.

\section{Video representation}
\label{sec:video_rep}
\begin{figure}[t]
\centering
\includegraphics[scale=0.2]{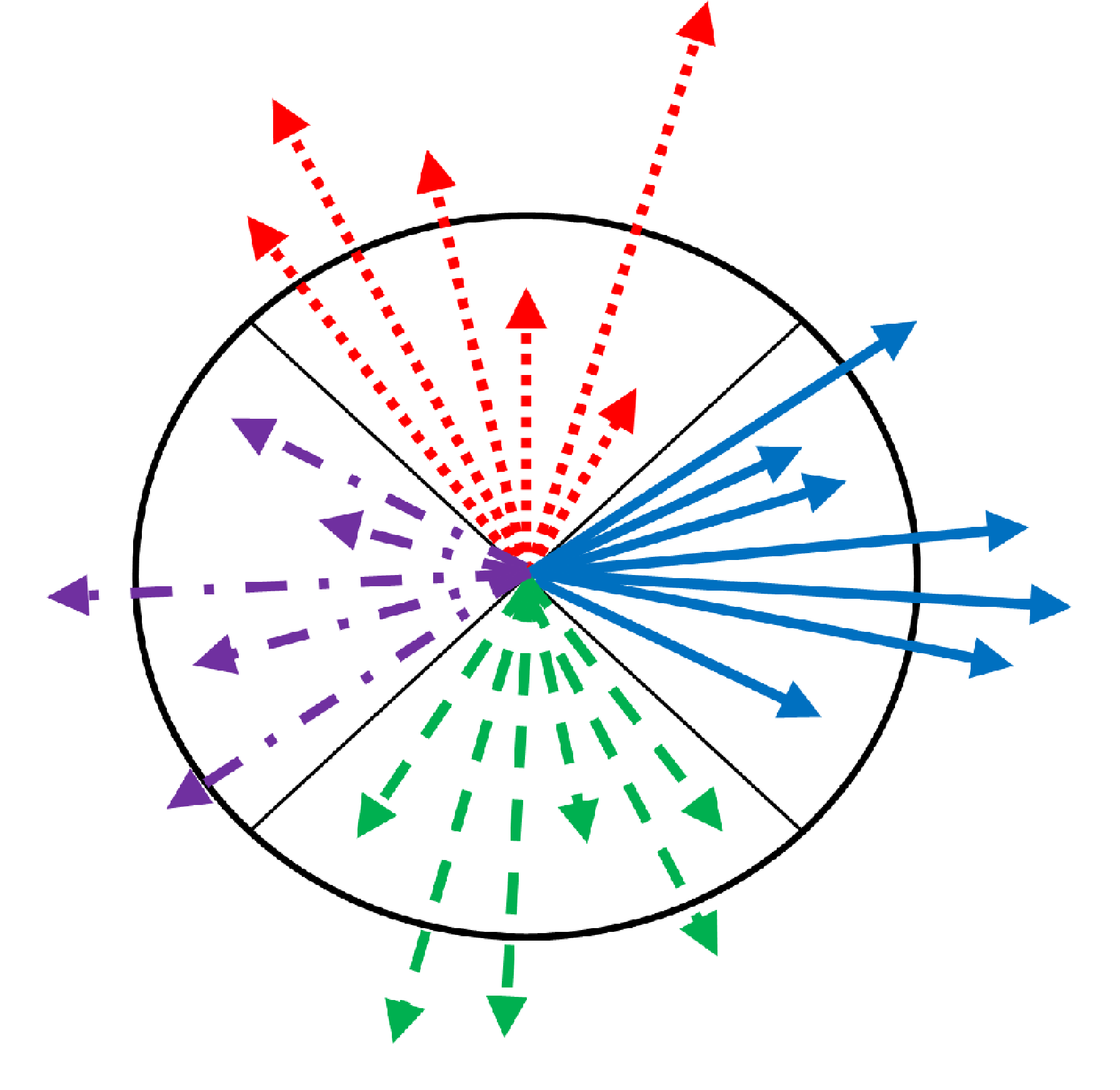}
\caption{Quantisation of motion directions. Optical flow vectors are quantised into the four directions --- up, right, down and left. The vectors of the same category have the same colour on the figure.}
\label{fig:word_formation}
\end{figure}

In order to apply the topic modeling approach to video processing it is required to define visual words and visual documents. In this paper a visual word is defined as a quantised local motion measured by an optical flow~\cite{Horn1981}. The optical flow vector is discretised spatially by averaging among $N \times N$ pixels. The direction of the average optical flow vector is further quantised into the four main categories --- up, right, down and left (Figure~\ref{fig:word_formation}). The location of the averaged optical flow vector and its categorised direction together form a visual word. 

The whole video sequence is divided into non-overlapping clips. Each clip is a visual document. The document consists of all the visual words extracted from the frames that form the corresponding clip. 

Topics in topic modeling are defined as distributions over words. They indicate which words appear together. In the video processing applications topics are distributions over visual words. As visual words represent local motions, topics indicate the set of local motions that frequently appear together. They are usually called \textit{activities} or \textit{actions} (e.g.~\cite{Wang09, Hospedales2011,Varadarajan2009,Isupova2015}).

Once visual documents, words and topics are defined, the topic model for video processing can be formulated.  

\section{Proposed model}
\label{sec:proposed_model}
There is a sequence of documents $\mathbf{x}_{1:J} = \{\mathbf{x}_j\}_{j = 1 : J}$, where each document $\mathbf{x}_j$ consists of $N_j$ words $x_{ji}$: $\mathbf{x}_j = \{x_{ji}\}_{i = 1:N_j}$. It is assumed that words are generated from a set of hidden distributions $\{\boldsymbol\phi_k\}_{k = 1 : \infty}$, that are called \textit{topics} and documents are mixtures of this shared set of topics. The number of topics is not fixed. Moreover it is assumed that observing the infinite amount of data we can expect to have an infinite number of topics. 

\subsection{Hierarchical Dirichlet Process Topic Model}
This kind of mixture models with a potentially infinite number of mixture components can be modelled with the Hierarchical Dirichlet Process (HDP) \cite{Teh2012}. The HDP is a hierarchical extension of the Dirichlet process (DP), which is a distribution over random distributions \cite{Ferguson1973}. Each document~$\mathbf{x}_j$ is associated with a sample $G_j$ from a DP:
\begin{equation}
\label{eq:HDP_def_start}
G_j \sim \text{DP}(\alpha, G_0),
\end{equation}
where $\alpha$ is a concentration parameter, $G_0$ is a base measure. $G_j$ can be seen as a vector of mixture components weights, where the number of components is infinite. 

The base measure $G_0$ itself is a sample from another DP:
\begin{equation}
G_0 \sim \text{DP}(\gamma, H),
\end{equation}
with the concentration parameter $\gamma$ and the base measure $H$. This shared measure $G_0$ from a DP ensures that the documents will have the same set of topics but with different weights. Indeed, $G_0$ is almost surely discrete \cite{Ferguson1973}, concentrating its mass on the atoms $\boldsymbol\phi_k$ drawn from $H$. Therefore, $G_j$ picks the mixture components from this set of atoms. 

A topic, that is an atom $\boldsymbol\phi_k$, is often modelled as the multinomial distribution with a probability $\phi_{wk}$ of choosing a word $w$ \cite{Hofmann99, Blei03LDA}. The base measure $H$ is therefore chosen as the conjugate Dirichlet distribution, usually a symmetric one. Let $\boldsymbol\eta = [\eta, \dotsc, \eta]$ denote a parameter of this Dirichlet distribution.

The document $j$ is formed by repeating the procedure of drawing a topic from the mixture:
\begin{equation}
\theta_{ji} \sim G_j
\end{equation}
and drawing a word from the chosen topic:
\begin{equation}
\label{HDP_def_finish}
x_{ji} \sim \text{Mult}(\theta_{ji})
\end{equation}
for every token $i$, where $\text{Mult}(\cdot)$ is the multinomial distribution.

\subsubsection{Chinese restaurant franchise}
There are several ways of the HDP representation (as well as the DP). In this paper the representation called Chinese restaurant franchise (CRF) is considered as it is used for the derivation of the Gibbs sampling inference scheme. In this metaphor, each document corresponds to a ``restaurant''; words correspond to ``customers'' of the restaurant. The words in the documents are grouped around ``tables''. Each table serves a ``dish'', which corresponds to a topic. The ``menu'' of dishes, i.e. the set of the topics, is shared among all the restaurants.  

Let $t_{ji}$ denote a table assignment for the token $i$ in the document $j$, $k_{jt}$ denote a topic assignment for the table $t$ in the document $j$. Let $n_{jt}$ denote the number of words assigned to the table $t$ in the document $j$ and $m_{jk}$ denote the number of tables in the document $j$ serving the topic $k$. The dots in subscripts mean marginalisation over the corresponding dimension, e.g. $m_{\cdot k}$ denotes the number of tables among all the documents serving the topic $k$, while $m_{j \cdot}$ denotes the total number of tables in the document $j$. Marginalisation over both dimensions $m_{\cdot\cdot}$ means the total number of tables in the dataset.

The generative process of a dataset is as follows. A new token comes to the document $j$ and chooses one of the occupied tables with a probability proportional to a number of words $n_{jt}$ assigned to this table, or the new token starts a new table with a probability proportional to $\alpha$:
\begin{equation}
\label{eq:hdp_t}
p(t_{ji} = t | t_{j1}, \dotsc, t_{j i-1}, \alpha) = 
\begin{cases}
\dfrac{n_{jt}}{i - 1 + \alpha}, \text{ if } t = 1 : m_{j\cdot}; \\[1.5mm]
\dfrac{\alpha}{i - 1 + \alpha}, \text{ if } t = t^{\text{new}}.
\end{cases}
\end{equation}

If the token starts a new table it chooses one of the used topics with a probability proportional to a number of tables $m_{\cdot k}$ serving this topic among all the documents, or the token chooses a new topic, sampling it from the base measure $H$, with a probability proportional to $\gamma$:
\begin{equation}
\label{eq:hdp_k}
p(k_{j t^{\text{new}}} = k | k_{11}, \dotsc, k_{jt-1}, \gamma) =
\begin{cases} 
\dfrac{m_{\cdot k}}{m_{\cdot \cdot} + \gamma}, \text{ if } k = 1 : K; \\[1.5mm]
\dfrac{\gamma}{m_{\cdot \cdot} + \gamma}, \text{ if } k = k^{\text{new}},
\end{cases}
\end{equation} 
where $K$ is a number of topics used so far. 

Once the token is assigned to the table $t_{ji}$ with the topic $k_{j t_{ji}}$, the word $x_{j i}$ for this token is sampled from this topic:
\begin{equation}
\label{eq:hdp_x}
x_{j t} \sim \text{Mult}(\boldsymbol\phi_{k_{j t_{j i}}})
\end{equation}

The correspondence between two representations of the HDP (\ref{eq:HDP_def_start}) -- (\ref{HDP_def_finish}) and (\ref{eq:hdp_t}) -- (\ref{eq:dynHDP_x}) is based on the following equality:  $\theta_{j i} = \boldsymbol\phi_{k_{j t_{j i}}}$.

\subsection{Dynamic Hierarchical Dirichlet Process Topic Model}

In the HDP exchangeability of documents and words is assumed which means that the joint probability of the data is independent of the order of the documents and the words in the documents. However, in the video processing applications this assumption may be invalid. While the words inside the documents are still exchangeable, the documents themselves are not. All actions and motions in the real life last for some time, and it is expected that the topic mixture in the current document is similar to the topic mixture in the previous document. Some topics may appear and disappear but the core structure of the mixture components weights only slightly changes from document to document. 

We propose the dynamic extension of the HDP topic model to take into account this intuition. In this model the probability of the topic $k$ explicitly depends on the usage of this topic in the current and previous documents $m_{j k} + m_{j-1 k}$, therefore the topic distribution in the current document would be similar to the topic distribution in the previous document. The topic probability still depends on the number of tables serving this topic in the whole dataset $m_{\cdot k}$, but this number is weighted by a non-negative value $\delta$, which is a parameter of the model. As in the previous case, it is possible to sample a new topic from the base measure $H$. 

The generative process can be then formulated as follows. A new token comes to a document and, as before, chooses one of the occupied tables $t$ with a probability proportional to the number of words $n_{j t}$ already assigned to it, or it starts a new table with a probability proportional to the parameter $\alpha$: 
\begin{equation}
\label{eq:dynHDP_t}
p(t_{ji} = t | t_{j 1}, \dotsc, t_{j i-1}, \alpha) = 
\begin{cases}
\dfrac{n_{j t}}{i - 1 + \alpha}, \text{ if } t = 1 : m_{j \cdot}; \\[1.5mm]
\dfrac{\alpha}{i - 1 + \alpha}, \text{ if } t = t^{\text{new}}.
\end{cases}
\end{equation}

If the token starts a new table, it chooses a topic for it. One of the used topics $k$ is chosen with a probability proportional to the sum of the number of tables having this topic in the current and previous documents $m_{jk} + m_{j-1 k}$ and the weighted number of tables among all the documents, which serve this topic, $\delta \, m_{\cdot k}$. A new topic can be chosen for the table $t$ with a probability proportional to the parameter $\gamma$:
\begin{align}
&p(k_{jt} = k | k_{11}, \dotsc, k_{jt-1}, \gamma) = \nonumber\\
\label{eq:dynHDP_k}
&\begin{cases}
\dfrac{m_{jk} + m_{j-1 k} + \delta m_{\cdot k}}{m_{j \cdot} + m_{j-1 \cdot} + \delta m_{\cdot \cdot} + \gamma}, \text{ if } k = 1:K;\\[1.5mm]
\dfrac{\gamma}{m_{j\cdot} + m_{j-1 \cdot} + \delta m_{\cdot \cdot} + \gamma}, \text{ it } k = k^{\text{new}}.
\end{cases}
\end{align}

Finally, the word $x_{j i}$ is sampled for the token $i$ in the document $j$, assigned to the table $t_{j i} = t$, which serves the topic $k_{j t} = k$. The word is sampled from the corresponding topic $k$:
\begin{equation}
\label{eq:dynHDP_x}
x_{j i} \sim \text{Mult}(\boldsymbol\phi_k).
\end{equation}

\section{Inference}
\label{sec:inference}
Standard inference algorithms process an entire dataset. For large or stream datasets this batch set up is computationally intractable. Online algorithms process data in a sequential manner, one data point at a time, incrementally updating the variables, corresponding to the whole dataset. It allows to save memory space and reduce the computational time. In this paper a combination of offline batch and online inference is proposed and this section describes it in details.   

The Gibbs sampling scheme is used \cite{Geman1984}. The inference procedure consists of two parts. Firstly, the traditional batch set up of the Gibbs sampling is applied to the training set of the documents. Then an online set up of the inference is applied for the testing documents. This means that the information about a testing document is incrementally added to the model, not requiring to process the training documents again.

In the Gibbs sampling inference scheme the hidden variables $\mathbf{t} = \{t_{j i}\}_{j = 1 : J, i = 1 : N_j}$ and $\mathbf{k} = \{k_{j t}\}_{j = 1 : J, t = 1 : m_{j \cdot}}$ are sampled from their conditional distributions. In the Gibbs sampler for the HDP model exchangeability of documents and words is used by treating the current variable $t_{j i}$ as the table assignment for the last token in the last document and $k_{j t}$ as the topic assignment for the last table in the last document. There is no exchangeability of documents in the proposed model, but words inside a document are still exchangeable. Therefore, the variable $t_{j i}$ can be treated as the table assignment for the last token in the current document $j$, and the variable $k_{j t}$ can be treated as the topic assignment for the last table in the current document $j$. The documents are processed in the order they appear in the dataset.

The following notation is used below. 
Let $V$ denote the size of the words vocabulary, $\mathbf{t}_{j_1:j_2} = \{t_{j i}\}_{j = j_1:j_2, i = 1:N_j}$ is the set of the table assignments for all the tokens in the documents from $j_1$ to $j_2$. Let $\mathbf{k}_{j_1:j_2} = \{k_{j t}\}_{j = j_1 : j_2, t = 1 : m_{j \cdot}}$ and $\mathbf{x}_{j_1:j_2} = \{\mathbf{x}_{j}\}_{j = j_1:j_2}$ denote the corresponding sets for the topic assignments and the observed data. 
Let $m_{j_1:j_2 \, k}$ denote the number of tables having the topic $k$ in the documents from $j_1$ to $j_2$.
Let also $\mathbf{x}_{jt} = \{x_{ji}\}_{i = 1 : N_j}$ denote the words assigned to the table $t$ in the document $j$.

Let $l_{w k}$ denote the number of times the word $w$ is associated with the topic $k$, $l_{\cdot k}$ denote the number of tokens associated with the topic $k$: $l_{\cdot k} = \sum\limits_w l_{w k}$, regardless the word assignments. The notation $l_{wk}^{j_1:j_2}$ is used for the number of times the word $w$ associated with the topic $k$ in the documents from $j_1$ to $j_2$.

The superscript $-ji$ indicates the corresponding variable without considering the token $i$ in the document $j$, e.g. the set variable $\mathbf{t}^{-ji} = \mathbf{t} \setminus \{t_{j i}\}$ or the count $n_{j t}^{-ji}$ is the number of words, assigned the table $t$ in the document $j$, excluding the word for the token $i$. Similarly, the superscript $-jt$ means the corresponding variable without considering the table $t$ in the document $j$. 

\subsection{Batch Gibbs sampling}
\subsubsection{Sampling topic assignment $k_{jt}$}
The topic assignment~$k_{jt}$ for the table~$t$ in the document~$j$ is sampled from the conditional distribution given the observed data~$\mathbf{x}$ and all the other hidden variables, i.e. the table assignments for all the tokens~$\mathbf{t}$ and the topic assignments for all the other tables~$\mathbf{k}^{-jt}$:
\begin{align}
&p(k_{jt} = k | \mathbf{x}, \mathbf{t}, \mathbf{k}^{-jt}) \propto \nonumber\\
\label{eq:k_sampling_start}
&p(\mathbf{x}_{jt} | k_{jt} = k, \mathbf{k}^{-jt}, \mathbf{t}, \mathbf{x}^{-jt})\, p(k_{jt} = k | \mathbf{k}^{-jt}).
\end{align}

The likelihood term $p(\mathbf{x}_{jt} | k_{jt} = k, \mathbf{k}^{-jt}, \mathbf{t}, \mathbf{x}^{-jt})$ can be computed by integrating out the distribution $\boldsymbol\phi_k$:
\begin{align}
&f_k^{-jt}(\mathbf{x}_{jt}) \stackrel{\text{def}}{=}  p(\mathbf{x}_{jt} | k_{jt} = k, \mathbf{k}^{-jt}, \mathbf{t}, \mathbf{x}^{-jt}) = \nonumber\\
&\int p(\mathbf{x}_{jt} | \boldsymbol\phi_k)\, p(\boldsymbol\phi_k | \mathbf{k}^{-{jt}}, \mathbf{t}, \mathbf{x}^{-jt}) \mathrm{d}\boldsymbol\phi_k = \nonumber\\
\label{eq:k_sampling_likelihood_start}
&\dfrac{\prod_w \Gamma(l_{w k} + \eta)}{\vphantom{\prod_{w}}\Gamma(l_{\cdot k} + V \eta)} \dfrac{\vphantom{\prod_w}\Gamma(l_{\cdot k}^{-jt} + V \eta)}{\prod_{w} \Gamma(l_{w k}^{-jt} + \eta)},
\end{align}
where $\Gamma(\cdot)$ is the gamma-function. In the case when $k$ is a new topic ($k = k^{\text{new}}$) the integration is done over the prior distribution for $\boldsymbol\phi_{k^{\text{new}}}$. The obtained likelihood term~(\ref{eq:k_sampling_likelihood_start}) is then:
\begin{equation}
\label{eq:k_sampling_likelihood_new}
f_{k^{\text{new}}}^{-jt}(\mathbf{x}_{jt}) = \dfrac{\prod_w \Gamma(l_{w k^{\text{new}}} + \eta)}{\Gamma(l_{\cdot k^{\text{new}}} + V \eta)} \dfrac{\vphantom{\prod_w}\Gamma(V\eta)}{(\Gamma(\eta))^{V}}.
\end{equation} 

The second multiplier in (\ref{eq:k_sampling_start}) $p(k_{jt} = k | \mathbf{k}^{-jt})$ can be further factorised as:
\begin{align}
&p(k_{jt} = k | \mathbf{k}^{-jt}) \propto \nonumber\\
\label{eq:k_sampling_topic_term}
&p(\mathbf{k}_{j+1:J} | \mathbf{k}_{1:j}^{-jt}, k_{jt} = k)\, p(k_{jt} = k | \mathbf{k}_{1:j}^{-jt}).
\end{align}

The first term in (\ref{eq:k_sampling_topic_term}) is the probability of the topic assignments for all the tables in the next documents depending on the change of the topic assignment for the table $t$ in the document $j$. Consider the topic assignments in the document $j+1$ firstly. From (\ref{eq:dynHDP_k}) it is:
\begin{align}
&g_{k}^{-jt}(\mathbf{k}_{j+1}) \stackrel{\text{def}}{=} p(\mathbf{k}_{j+1} | \mathbf{k}_{1:j}^{-jt}, k_{jt} = k) = \nonumber\\
&\dfrac{\gamma^{|K_{j+1}^{\text{born}}|} \prod_{s \in K_{j+1}^{\text{born}}} (m_{j+1 s} - 1)! (1 + \delta)^{m_{j+1 s} - 1}}{\prod_{n= 1}^{m_{j+1 \cdot}} (m_{j \cdot} + n - 1 + \delta(m_{1:j \,\cdot} + n - 1) + \gamma)} \, \times \nonumber\\
&\prod_{s\not\in K_{j+1}^{\text{born}}} \prod_{n = 1}^{m_{j+1\, s}} (m_{j s}^{-jt\rightarrow k} + n - 1 + \delta (m_{1:j\, s}^{-jt\rightarrow k} + n - 1))\propto \nonumber\\
\label{eq:k_sampling_next_doc_topics}
&\prod_{s\not\in K_{j+1}^{\text{born}}} \prod_{n = 1}^{m_{j+1\, s}} (m_{j s}^{-jt\rightarrow k} + n - 1 + \delta (m_{1:j\, s}^{-jt\rightarrow k} + n - 1)),
\end{align}
where the sign of proportionality is used w.r.t. $k_{jt}$, $K_{j+1}^{\text{born}}$ is the set of the topics that firstly appear in the document $j+1$, the superscript $-jt\rightarrow k$ means that $k_{jt}$ is set to $k$ for the corresponding counts, $|\cdot|$ is the cardinality of the set. The similar probabilities of the topic assignments for all the next documents $j' = j+2 : J$ depend on $k$ only in the term $m_{1:j'-1 \, \cdot}^{-jt\rightarrow k}$. It is assumed that the influence of $k$ on these probabilities is not significant and the first term in (\ref{eq:k_sampling_topic_term}) is approximated by the probability of the topic assignments in the document $j+1$ (\ref{eq:k_sampling_next_doc_topics}) only:
\begin{equation}
\label{eq:k_sampling_next_approx}
p(\mathbf{k}_{j+1:J} | \mathbf{k}_{1:j}^{-jt}, k_{jt} = k) \approx g_k^{-jt}(\mathbf{k}_{j+1}).
\end{equation}

The second term in (\ref{eq:k_sampling_topic_term}) is the prior for $k_{jt}$:
\begin{align}
&p(k_{jt} = k | \mathbf{k}_{1:j}^{-jt}) \propto \nonumber\\
\label{eq:k_sampling_prior}
&\begin{cases}
m_{jk}^{-{jt}} + m_{j-1 k} + \delta m_{1:j\,k}^{-jt}, \text{ if } k = 1: K;\\
\gamma, \text{ if } k = k^{\text{new}}.
\end{cases}
\end{align}

As a result, (\ref{eq:k_sampling_topic_term}) is computed as follows:
\begin{align}
&p(k_{jt} = k | \mathbf{k}^{-jt}) \propto \nonumber\\
\label{eq:k_cond}
&\begin{cases}
g_k^{-jt}(\mathbf{k}_{j+1}) (m_{jt} + m_{j-1 k} + \delta m_{1:j \, k}), \text{ if } k = 1 : K;\\
g_{k^{\text{new}}}^{-jt}(\mathbf{k}_{j+1}) \gamma, \text{ if } k = k^{\text{new}}.
\end{cases}
\end{align}

Combining~(\ref{eq:k_sampling_likelihood_start})~--~(\ref{eq:k_sampling_likelihood_new}) and~(\ref{eq:k_cond}) the topic assignment sampling distribution can be expressed as:
\begin{equation}
\label{eq:sample_k}
p(k_{jt} = k | \mathbf{x}, \mathbf{t}, \mathbf{k}^{-jt}) \propto f_k^{-jt}(\mathbf{x}_{jt})\, p(k_{jt} = k | \mathbf{k}^{-jt}).
\end{equation}

\subsubsection{Sampling $t_{j i}$} 
The table assignment $t_{j i}$ for the token $i$ in the document $j$ is sampled from the conditional distribution given the observed data $\mathbf{x}$ and all the other hidden variables, i.e. the topic assignments for all the tables $\mathbf{k}$ and the table assignments for all the other tokens $\mathbf{t}^{-ji}$:
\begin{align}
\label{eq:t_sampling_start}
&p(t_{j i} = t | \mathbf{x}, \mathbf{k}, \mathbf{t}^{-ji}) \propto \nonumber\\
&p(x_{j i} | \mathbf{t}^{-ji}, t_{j i} = t, \mathbf{x}^{-ji}, \mathbf{k})\, p(t_{j i} = t | \mathbf{t}^{-ji})
\end{align}

The first term in (\ref{eq:t_sampling_start}) is the likelihood of the word $x_{j i}$. It changes depending on whether $t$ is one of the previously used table or it is a new table. For the case when $t$ is the table which is already used the likelihood is:
\begin{equation}
f_{k_{jt}}^{-ji}(x_{ji}) = p(x_{ji} | t_{ji} = t, \mathbf{t}^{-ji}, \mathbf{k}, \mathbf{x}^{-ji}) = \dfrac{l_{x_{ji} \, k_{jt}} + \eta}{l_{\cdot \, k_{jt}} + V\eta}
\end{equation}

Consider now the case when $t_{j i} = t^{\text{new}}$, i.e. the likelihood of the word $x_{j i}$ being assigned to a new table. This likelihood can be found by integrating out the possible topic assignments $k_{j t^{\text{new}}}$ for this table:
\begin{align}
\label{eq:t_sampling_new_table_likelihood_start}
&r_{t^{\text{new}}}(x_{j i}) \stackrel{\text{def}}{=}
p(x_{j i} | \mathbf{t}^{-ji}, t_{j i} = t^{\text{new}}, \mathbf{x}^{-ji}, \mathbf{k}) = \nonumber\\
&\sum\limits_{k = 1}^K f_{k}^{-ji}(x_{j i})\, p(k_{j t^{\text{new}}} = k | \mathbf{k}) + \nonumber\\
&f_{k^{\text{new}}}^{-ji}(x_{j i})\, p(k_{j t^{\text{new}}} = k^{\text{new}} | \mathbf{k}),
\end{align}
where $p(k_{j t^{\text{new}}} = k | \mathbf{k})$ is as (\ref{eq:k_cond}).


The second term in~(\ref{eq:t_sampling_start}) is the prior for~$t_{ji}$:
\begin{equation}
p(t_{j i} = t | \mathbf{t}^{-ji}) \propto 
\begin{cases}
n_{jt}, \text{ if } t = 1 : m_{j \cdot}; \\
\alpha, \text{ if } t = t^{\text{new}}.
\end{cases}
\end{equation}

Then the conditional distribution for sampling a table assignment~$t_{ji}$ is:
\begin{align}
&p(t_{j i} = t | \mathbf{x}, \mathbf{k}, \mathbf{t}^{-ji}) \propto \nonumber\\
\label{eq:sample_t}
&\begin{cases}
f_{k_{jt}}^{-ji}(x_{ji}) n_{jt}, \text{ if } t = 1 : m_{j \cdot}; \\
r_{t^{\text{new}}}(x_{j i}) \alpha, \text{ if } t = t^{\text{new}}.
\end{cases}
\end{align}

If a new table is sampled, then a topic for it is sampled from~(\ref{eq:sample_k}).

\subsection{Online inference}
In online or distributed implementations of inference algorithms in topic modeling the idea is to separate global variables, i.e. those that depend on the whole set of data, and local variables, i.e. those that depend only on the current document~\cite{Vorontsov2015bigartm, Asuncion2009, Wang2011online}. 

For the proposed dynamic HDP model the global variables are the distributions $\boldsymbol\phi_k$, which are approximated by the counts $l_{w k}$, and the global topic popularity, which is estimated by the counts $m_{\cdot k}$. Note, that the relative relationship between counts is important, rather than the absolute values of the counts. The local variables are the topic mixture weights for each document, governed by the counts $m_{jk}$. The training dataset is assumed to be large enough such that the global variables are well estimated by the counts available during the training stage and a new document can only slightly change the obtained ratios of the counts. 

Following this assumption, the learning procedure is organised as follows. The batch Gibbs sampler is run for the training set of the documents. After this training stage the global counts $l_{wk}$ and $m_{\cdot k}$ for all $w$ and $k$ are stored and used for the online inference of the testing documents. For each testing document the online Gibbs sampler is run to sample table assignments and topic assignments for this document only. The online Gibbs sampler updates the local counts $m_{jk}$. After the Gibbs sampler converges, the global counts $l_{wk}$ and $m_{\cdot k}$ are updated with the information obtained by the new document. 

The equations for the online version of the Gibbs sampler slightly differ from the batch ones~(\ref{eq:sample_k}) and~(\ref{eq:sample_t}). Namely, the conditional probability $p(k_{jt} = k | \mathbf{k}^{-jt})$ in the topic assignment sampling distribution (\ref{eq:sample_k}) differs from (\ref{eq:k_sampling_topic_term}). As next documents are not observed during processing the current document, this probability consists only of the prior term $p(k_{jt} = k | \mathbf{k}_{1:j}^{-jt})$:
\begin{align}
&p_{\text{online}}(k_{jt} = k | \mathbf{k}^{-jt}) = \nonumber \\
&\begin{cases}
m_{jk}^{-{jt}} + m_{j-1 k} + \delta m_{1:j\,k}^{-jt}, \text{ if } k = 1: K;\\
\gamma, \text{ if } k = k^{\text{new}}.
\end{cases}
\end{align}
Substituting this expression into~(\ref{eq:sample_k}) the obtained sampling distribution for the topic assignment in the online Gibbs sampler is:
\begin{align}
&p_{\text{online}}(k_{jt} = k | \mathbf{x}, \mathbf{t}, \mathbf{k}^{-jt}) \propto \nonumber\\
&\begin{cases}
f_k^{-jt}(\mathbf{x}_{jt}) (m_{jt} + m_{j-1 k} + \delta m_{1:j \, k}), \text{ if } k = 1 : K;\\
f_{k^{\text{new}}}^{-jt}(\mathbf{x}_{jt}) \gamma, \text{ if } k = k^{\text{new}}.
\end{cases}
\end{align}

The updating distribution for the topic assignment in the online Gibbs sampler remains the same as in the batch version~(\ref{eq:sample_t}).

\section{Abnormality detection}
\label{sec:abnormality}
Topic models provide a probabilistic framework for abnormality detection. Under this framework the abnormality measure is the likelihood of data. The low value of the likelihood means the built model cannot explain the current observation, i.e. there is something atypical in the observation, which is not fitted to the typical motion patterns, learnt by the model. 

From the Gibbs sampler we have estimates of the distributions~$\boldsymbol\phi_k$ and posterior samples of the table and topic assignments. This information can be used to estimate the predictive likelihood of a new clip. The predictive likelihood, normalised by the length~$N_j$ of the clip in terms of visual words, is used as an abnormality measure in this paper. 

The predictive likelihood is estimated via a harmonic mean~\cite{Griffiths2004}, as it allows to use the information from the posterior samples:
\begin{align}
&p(\mathbf{x}_j | \mathbf{x}_{1:j-1}) = \nonumber\\
&\left(\sum\limits_{\mathbf{t}_{1:j}, \mathbf{k}_{1:j}} \dfrac{p(\mathbf{t}_{1:j}, \mathbf{k}_{1:j} | \mathbf{x}_j, \mathbf{x}_{1:j-1})}{p(\mathbf{x}_j | \mathbf{t}_{1:j}, \mathbf{k}_{1:j}, \mathbf{x}_{1:j-1})}\right)^{-1} \approx\nonumber\\
&\left( \dfrac{1}{S} \sum\limits_{s = 1}^{S} \dfrac{1}{p(\mathbf{x}_{j} | \mathbf{t}_{1:j}^s, \mathbf{k}^s, \mathbf{x}_{1:j-1})} \right),
\end{align}
where $S$ is the number of the posterior samples, $\mathbf{t}_{1:j}^s$ and $\mathbf{k}_{1:j}^s$ are from the $s$-th posterior sample obtained by the Gibbs sampler, and 
\begin{align}
&p(\mathbf{x}_{j} | \mathbf{t}_{1:j}^s, \mathbf{k}^s, \mathbf{x}_{1:j-1}) = \nonumber\\
&\prod\limits_{k = 1}^K \dfrac{\prod_w \Gamma(l_{w k}^{1:j \,s } + \eta)}{\vphantom{\prod_w}\Gamma(l_{\cdot k}^{1:j\, s} + V\eta)} \dfrac{\vphantom{\prod_w}\Gamma(l_{\cdot k}^{1:j-1\, s} + V\eta)}{\prod_w \Gamma(l_{w k}^{1:j-1\, s} + \eta)}.
\end{align}
The superscript $s$ on the counts means these counts are from the $s$-th posterior sample.

The abnormality detection procedure is then as follows. The batch Gibbs sampler is run on the training dataset. Then for each clip from the testing dataset first the online Gibbs sampler is run to obtain the posterior samples of the hidden variables corresponding to the current clip. Afterwards the abnormality measure: 
\begin{equation}
a(\mathbf{x}_j) = \dfrac{1}{N_j} p(\mathbf{x}_j | \mathbf{x}_{1:j-1})
\end{equation}
is computed for the current clip. If the abnormality measure is below than some threshold, the clip is labelled as abnormal, otherwise as normal. And the next clip from the testing dataset is processed.   

\section{Experiments}
\label{sec:experiments}
\begin{figure}[t]
\centering
\includegraphics[width=0.95\columnwidth]{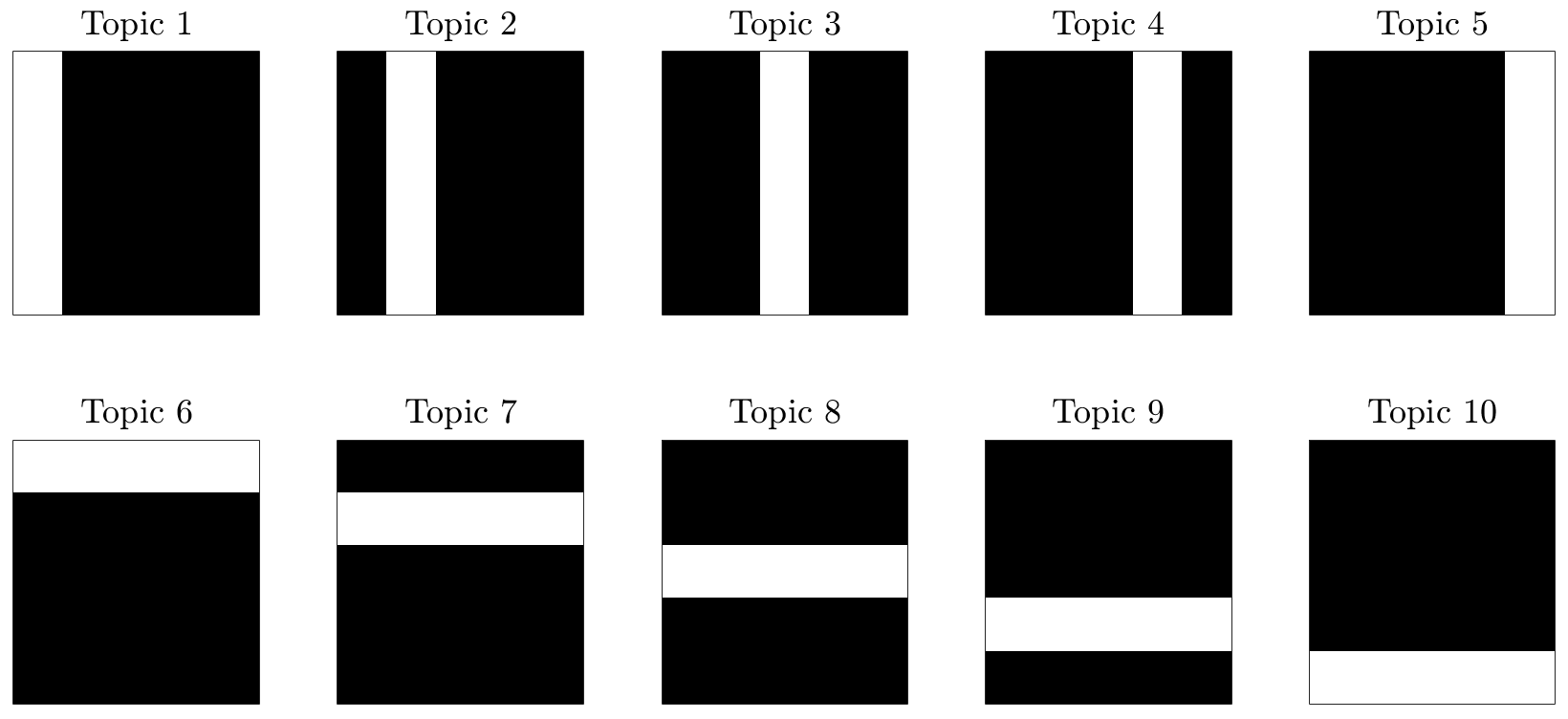}
\caption{Graphical representation of the topics in the synthetic dataset. There are $25$ words, organised into a $5 \times 5$ matrix, where a word corresponds to a cell in this matrix. The topics are represented as the coloured matrices, where the colour of the cell indicates the probability of the corresponding word in a given topic, the lighter the colour the higher the probability value.}
\label{fig:synthetic_topics}
\end{figure}

\begin{figure}[t]
\centering
\includegraphics[scale=0.4]{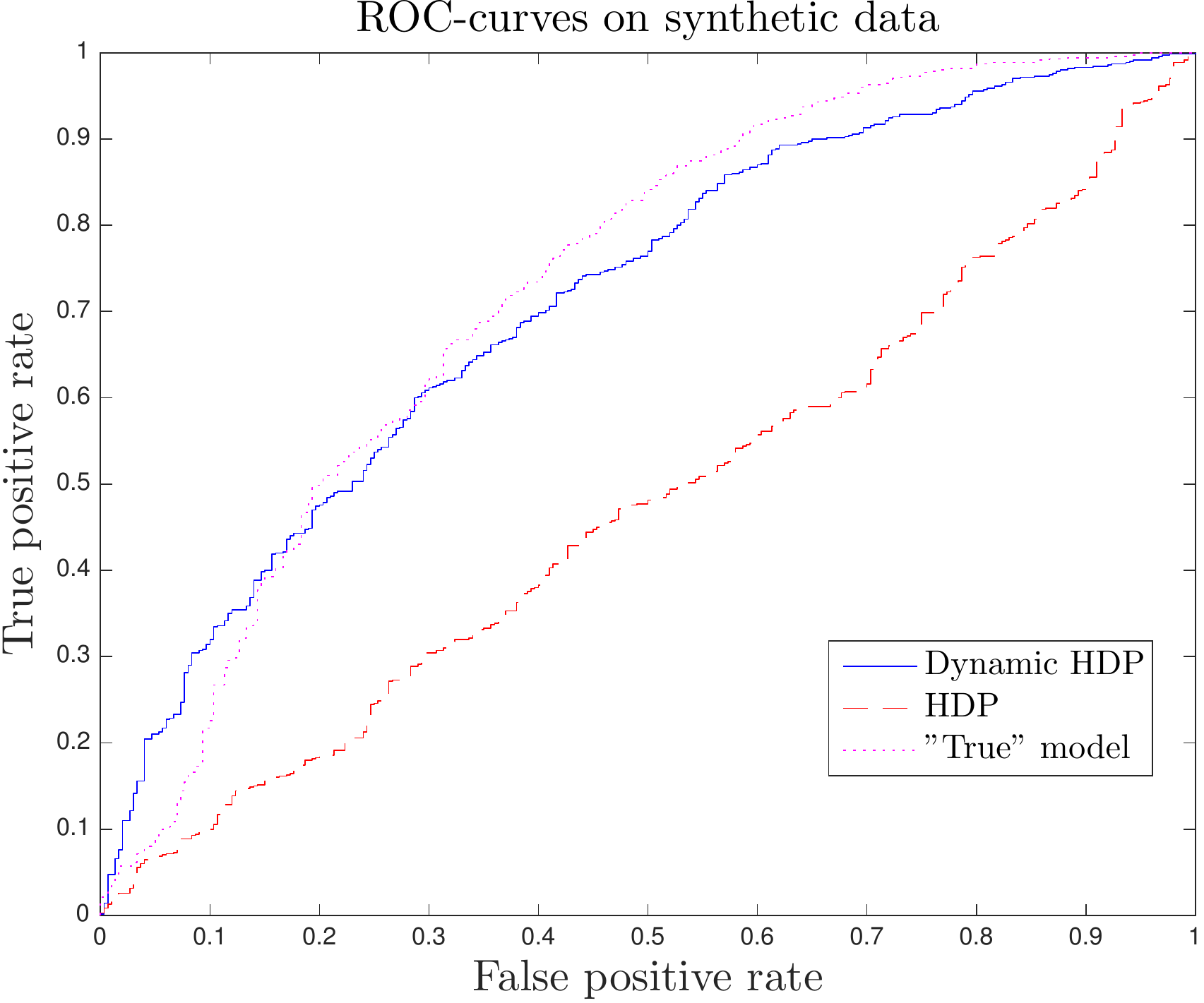}
\caption{The ROC-curves for the synthetic data obtained by both models. The ROC-curve, obtained by the likelihood, computed with the known true hidden variables, is labelled as a ``true'' model.}
\label{fig:synthetic_roc}
\end{figure}  

In this section the proposed method is applied to abnormality detection\footnote{The code is available on https://github.com/OlgaIsupova/dynamic-hdp}. The method is compared with the one, based on the HDP topic model, where for the HDP topic model the online version of the Gibbs sampler and the abnormality measure are derived similarly to the dynamic HDP (for the batch Gibbs sampler of the HDP topic model the implementation by Chong Wang is used\footnote{It is available on https://github.com/Blei-Lab/hdp}). Each of the algorithms has 5 runs with different initialisations to obtain 5 independent posterior samples. Both batch and online samplers are run for 1000 ``burn-in'' iterations.

The methods are compared on both synthetic and real data. The abnormality classification accuracy is used for the quantitative comparison of the methods. For computing classification accuracy the ground truth about abnormality should be provided. For the synthetic data the ground truth is known from the generation, for the test real data the clips are labelled manually as normal or abnormal. Note, the methods use only unlabelled data, labels are applied for performance measure.
  
In statistics the following measures are used for binary classification: \textit{true positive} (TP) is the number of observations which are correctly detected by an algorithm as positive, \textit{false negative} (FN) is the number of observations which are incorrectly detected as negative, \textit{true negative} (TN) is the number of observations which are correctly detected as negative, and \textit{false positive} {FP} is the number of observations which are incorrectly detected as positive~\cite{Murphy2012}. 

For the quantitative comparison the area (AUC) under the receiver operating characteristic (ROC) curve is used in this paper. The curve is built by plotting the true positive rate versus the false positive rate while the threshold varies. The true positive rate (TPR), also known as recall, is defined as:
\begin{equation}
TRP = \dfrac{TP}{TP + FN}.
\end{equation}
The false positive rate (FPR), also known as fall-out, is defined as:
\begin{equation}
FPR = \dfrac{FP}{FP + TN}.
\end{equation}

\begin{figure*}[ht]
\centering
\subfloat[]{\includegraphics[width = 0.47\columnwidth]{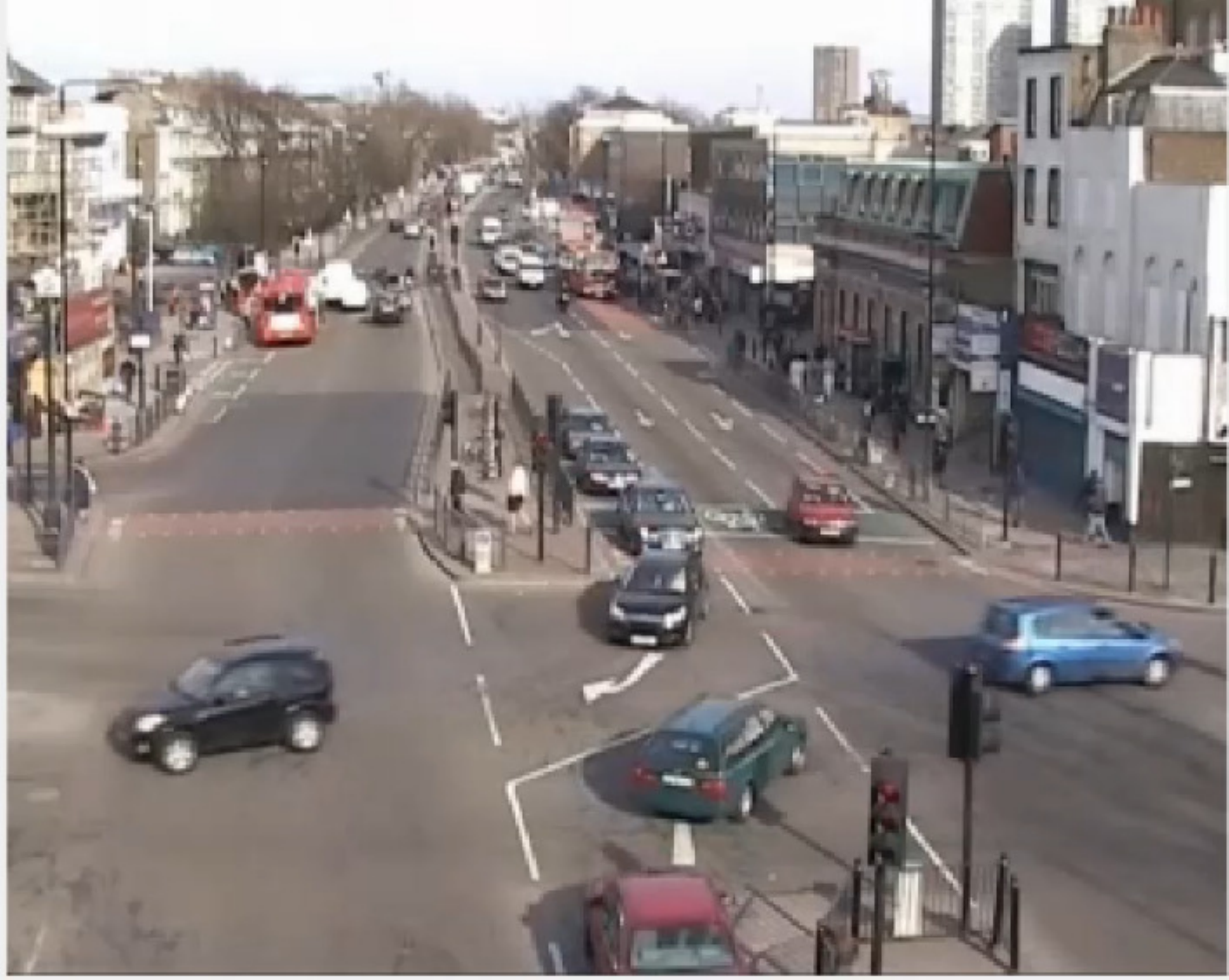}
\label{fig:qmul_normal}}%
\hfil
\subfloat[]{\includegraphics[width = 0.47\columnwidth]{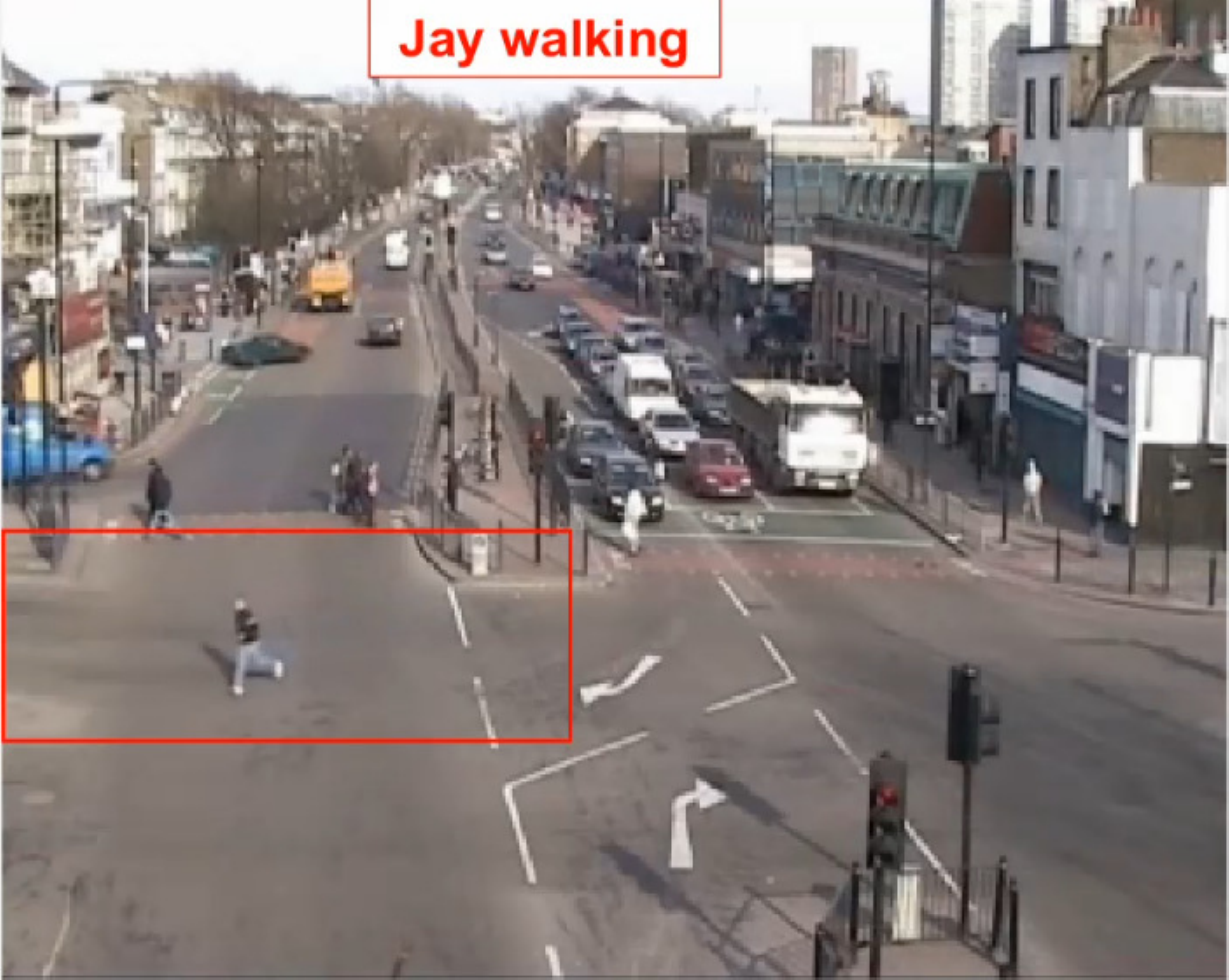}
\label{fig:qmul_jaywalking}}%
\hfil
\subfloat[]{\includegraphics[width = 0.47\columnwidth]{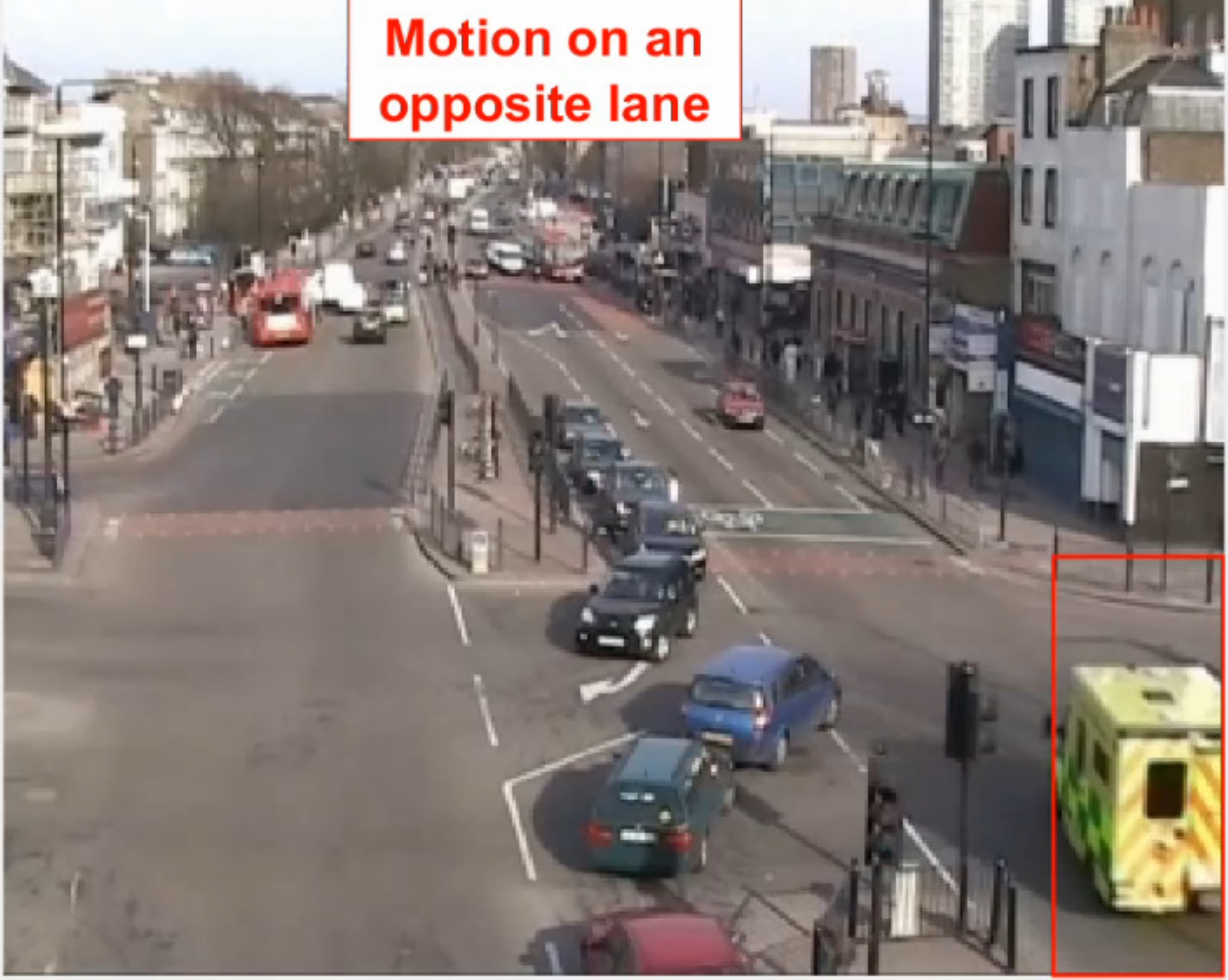}
\label{fig:qmul_wrong_direction}}%
\hfil
\subfloat[]{\includegraphics[width = 0.47\columnwidth]{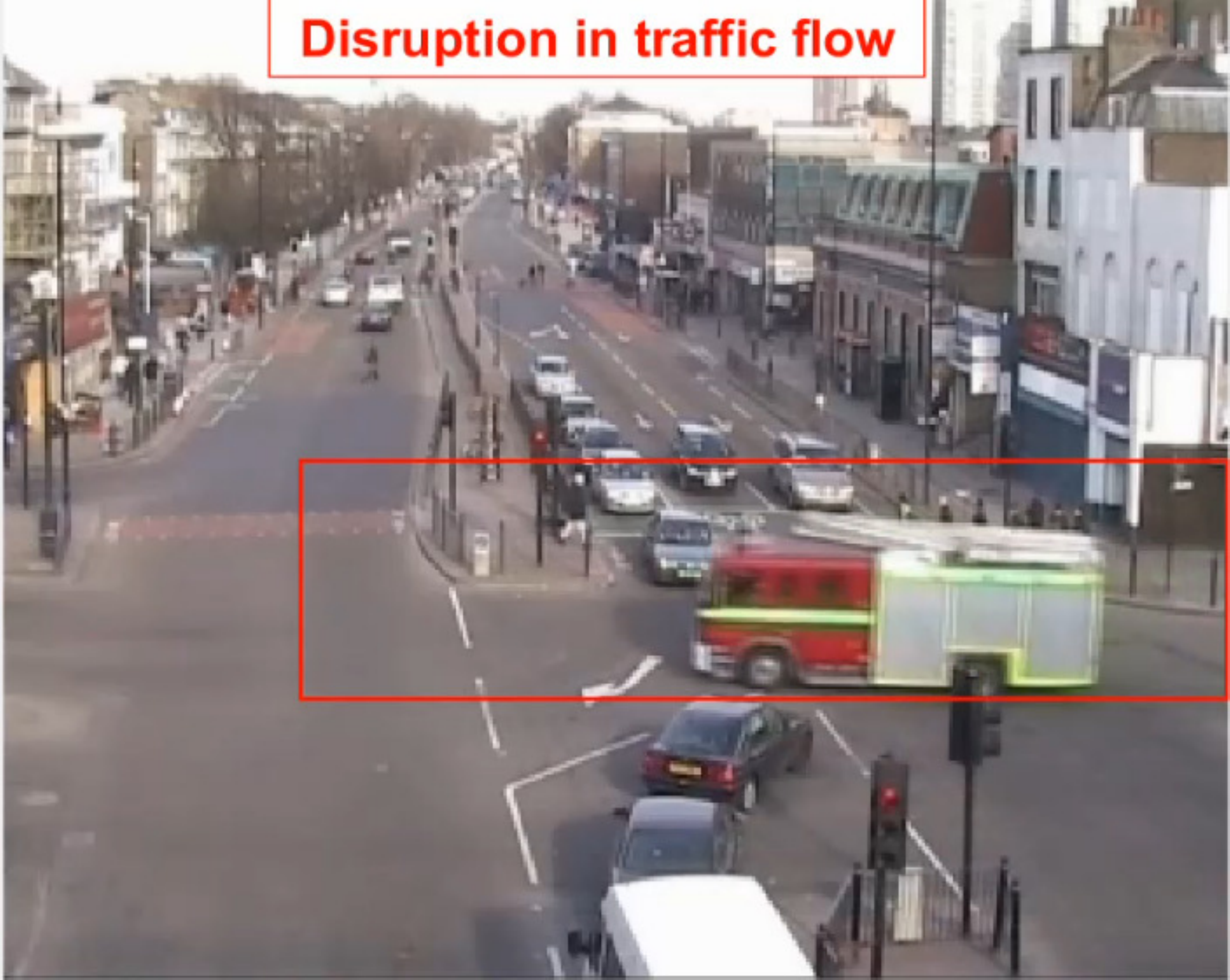}
\label{fig:qmul_disruption}}%
\caption{QMUL-junction dataset snapshots. \protect\subref{fig:qmul_normal} is an example of a normal motion, \protect\subref{fig:qmul_jaywalking} is an example of jay-walking abnormality, \protect\subref{fig:qmul_wrong_direction} is an example of a car moving on the wrong lane in the opposite to normal direction, \protect\subref{fig:qmul_disruption} is an example an emergency service car disrupting a normal traffic flow.}
\label{fig:qmul_samples}
\end{figure*}

\subsection{Synthetic data}
The popular ``bar'' data is used as a synthetic data (introduced in~\cite{Griffiths2004}). In this data the vocabulary consists of $V = 25$ words, organised into a $5 \times 5$ matrix. There are $10$ topics in total, the word distributions $\boldsymbol\phi_k$ of these topics form vertical and horizontal bars in the matrix (Figure~\ref{fig:synthetic_topics}). 

\begin{figure}[t]
\centering
\includegraphics[scale=0.4]{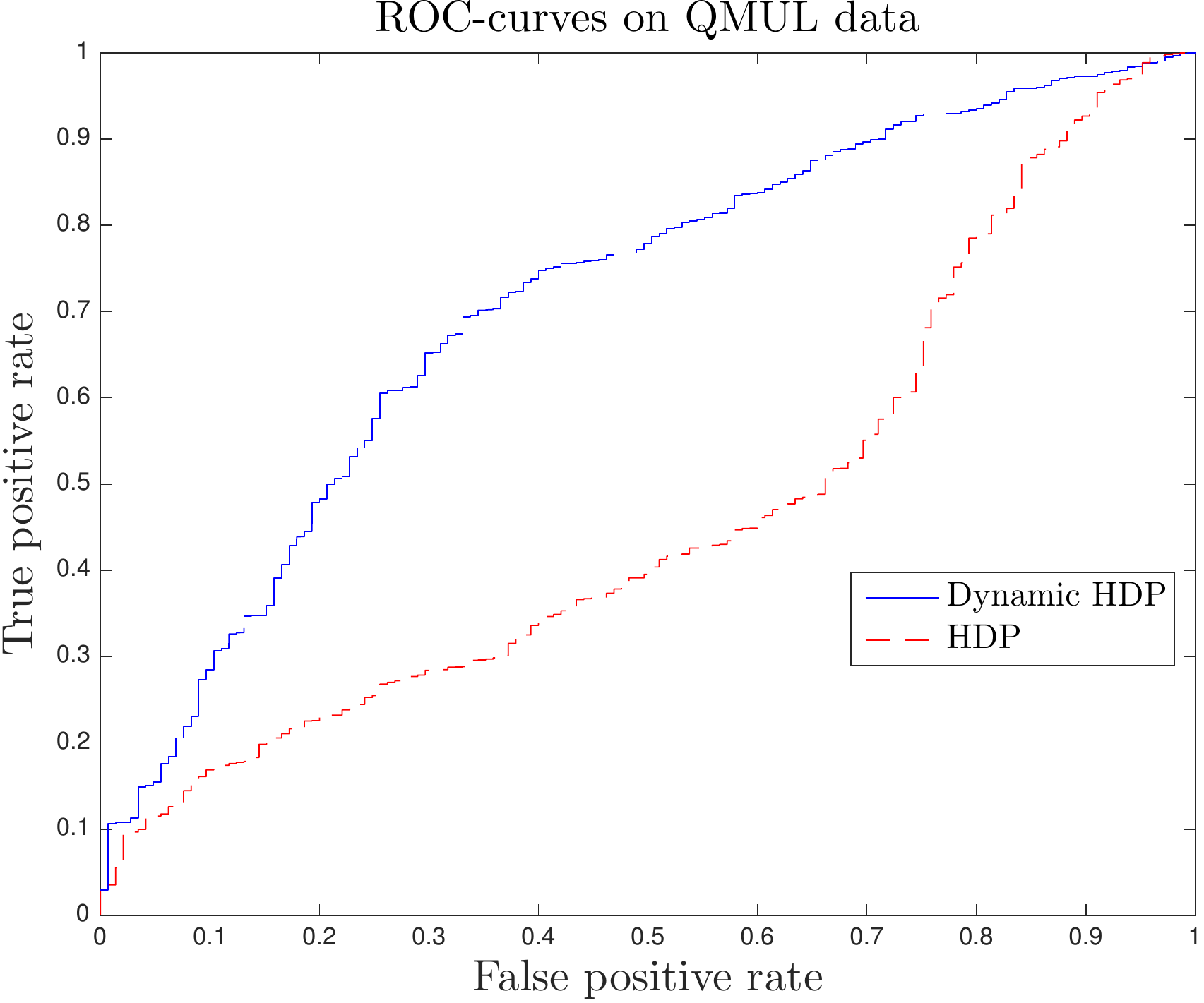}
\caption{The ROC-curves for the QMUL data.}
\label{fig:qmul_roc}
\end{figure}  


The training dataset consisting of $2000$ documents is generated from the proposed model~(\ref{eq:dynHDP_t})~--~(\ref{eq:dynHDP_x}), where $1\%$ noise is added to the distributions $\boldsymbol\phi_k$. Each of the documents has $20$ words. The hyperparameters are set to the following values for the generation: $\alpha = 1.5$, $\gamma = 2$, $\delta = 0.5$. 

Similarly, the testing dataset consisting of $1000$ documents is generated, but where $300$ random documents are generated as ``abnormal''. In the proposed model it is assumed that topic mixtures in neighbour documents are similar. Contrarily to this assumption topics for an abnormal document are chosen uniformly from the set of all the topics except those used in the previous document. 

The both algorithms are run for these datasets, computing the abnormality measure for all the testing documents. The hyperparameters $\alpha$, $\gamma$, $\delta$ are set to the same values as for the generation, $\eta = 0.2$ ($\eta$ is not used in the generation as the word distributions in topics are set manually).  

In Figure \ref{fig:synthetic_roc} the ROC-curves for the obtained abnormality measures are presented. There is also presented the ROC-curve for the ``true'' abnormality measure. The ``true'' abnormality measure is computed using the likelihood given the true distributions $\boldsymbol\phi_k$ and the true table and topic assignments $\mathbf{t}$ and $\mathbf{k}$, i.e. it corresponds to the model that can perfectly restore the latent variables. Table~\ref{tab:auc} contains the obtained AUC values. 

\begin{table}[t]
\renewcommand{\arraystretch}{1.3}
\caption{AUC results}
\label{tab:auc}
\centering
\begin{tabular}{|c|c|c|c|}
\hline
\textbf{Dataset} & \textbf{Dynamic HDP} & \textbf{HDP} & \textbf{``True'' model}\\
\hline
\hline
Synthetic & 0.7118 & 0.4751 & 0.7280\\
\hline
QMUL & 0.7100 & 0.4644 & ---\\
\hline
\end{tabular}
\end{table}

The results show that the proposed dynamic HDP can detect the simulated abnormalities and its performance is competitive to the ``true'' model. The original HDP method should not detect this kind of abnormalities, as they do not contradict to its generative model, it is confirmed by the experimental results.

\subsection{Real data}
The algorithms are applied to the QMUL-junction real data~\cite{Hospedales2011}. This is a 45-minutes video captured a road junction~(Figure~\ref{fig:qmul_normal}). The frame size is $360 \times 288$. The $8 \times 8$-pixel grid cells are used for spatial averaging of the optical flow. For the optical flow estimation the sparse pyramidal version of the Lucas-Kanade optical flow algorithm is used~\cite{Bouguet2001} (the implementation is available in the opencv library). The resulting vocabulary size is $V = 6480$. Non-overlapping clips, $1$-second length, are treated as visual documents. A 5-minute video sequence is used as a training dataset. 

The algorithms are run with the following hyperparameters: $\alpha = 1$, $\gamma = 1$, $\eta = 0.5$. The weight parameter $\delta$ for the dynamic HDP is set to $1$.

The data is manually labelled as normal/abnormal to measure classification accuracy, where abnormal event examples are jay-walking (Figure~\ref{fig:qmul_jaywalking}), driving wrong direction~(Figure~\ref{fig:qmul_wrong_direction}), disruption in traffic flow~(Figure~\ref{fig:qmul_disruption}).

The ROC-curves for the methods are presented in Figure~\ref{fig:qmul_roc}. The corresponding AUC values can be found in Table~\ref{tab:auc}. 
The proposed dynamic HDP method outperforms the other one. The provided results show that consideration of dynamics in a topic model may improve the classification results in abnormality detection.

\section{Conclusions}
\label{sec:conclusions}
In this paper a novel Bayesian nonparametric dynamic topic model is proposed, denoted as dynamic HDP. The Gibbs sampling scheme is applied for inference. The online set up for the inference is designed, allowing to incrementally train the model when the data is processed sequentially. The model is applied for abnormal behaviour detection in video. The abnormality decision rule is based on the predictive likelihood of the data that is developed in this paper. We show that the proposed method, based on the dynamic topic model, improves the classification performance in comparison to the method, based on the model without dynamics. We compare the proposed dynamic HDP method with the method based on the HDP, introduced in \cite{Teh2012}. The experiments both on synthetic and real data confirm the superiority of the proposed method.

\section*{Acknowledgments}

Olga Isupova and Lyudmila Mihaylova would like to thank the support from the EC Seventh Framework Programme [FP7 2013-2017] TRAcking in compleX sensor systems (TRAX) Grant agreement no.: 607400. Lyudmila Mihaylova acknowledges also the support from the UK Engineering and Physical Sciences Research Council (EPSRC) via the Bayesian Tracking and Reasoning over Time (BTaRoT) grant EP/K021516/1.

\bibliography{Biblist}

\end{document}